\documentclass[authoryear,preprint,12pt]{elsarticle}

\usepackage{microtype}          
\usepackage{xurl}               
\usepackage[unicode,hidelinks,breaklinks=true]{hyperref}
\makeatletter
\g@addto@macro\UrlBreaks{\do\/\do-\do\_\do.\do?\do&\do=\do:}
\makeatother

\usepackage{doi}

\usepackage{amssymb}
\usepackage{amsmath}
\usepackage[ruled,vlined]{algorithm2e}
\usepackage{booktabs}
\usepackage{mathtools}
\DeclarePairedDelimiter{\floor}{\lfloor}{\rfloor}

\journal{Machine Learning with Applications}

\begin{document}

\begin{frontmatter}



\title{ART: Adaptive Resampling-based Training for Difficulty-aware Classification}


\author[1]{Arjun Basandrai\fnref{equal}} \ead{arjun.basandrai2022@vitstudent.ac.in} 
\author[1]{Shourya Jain\fnref{equal}} \ead{shourya.jain2022@vitstudent.ac.in} 
\author[1]{Ilanthenral Kandasamy \corref{cor1}} \cortext[cor1] {Corresponding author} \ead{ilanthenral.k@vit.ac.in} 

\fntext[equal]{These authors contributed equally to this work.}

\affiliation[1]{organization={School of Computer Science and Engineering, Vellore Institute of Technology},
         addressline={Katpadi-Tiruvalam Road}, city={Vellore}, postcode={632014},             state={Tamil Nadu},country={India}}

\begin{abstract}
Traditional resampling methods for addressing class imbalance in supervised classification typically use fixed sampling distributions, either uniformly undersampling the majority class or oversampling the minority class. These static strategies fail to account for changes in class-wise learning difficulty during the training process. This paper proposes an Adaptive Resampling-based Training (ART) method that periodically updates the distribution of the training data based on the model's class-wise performance. Specifically, ART uses class-wise macro F1 scores computed at fixed intervals to determine the degree of resampling to perform.

In contrast to instance-level difficulty modeling, which can be noisy and overly sensitive to outliers, ART adapts at the class level using the defined performance metric. This allows the model to incrementally shift its attention towards underperforming classes in a way that better aligns with the optimization objective. 

Experimental results across diverse class-imbalanced benchmark datasets demonstrate that ART consistently outperforms both resampling-based and algorithm-level methods, including Synthetic Minority Oversampling Technique, nearmiss undersampling, and cost-sensitive learning on binary as well as multi-class classification tasks with varying degrees of imbalance. 

In most settings, these improvements are statistically significant. On tabular datasets, gains are significant under both paired t-tests and Wilcoxon signed-rank tests (p < 0.05), while performance on text and image tasks remains consistently favorable. ART improves macro F1 by an average of 2.64 percentage points across all tested tabular datasets. Unlike existing methods, ART consistently delivers the highest macro F1 score, making it a reliable and broadly effective choice for imbalanced classification problems.
\end{abstract}



\end{frontmatter}



\section{Introduction}
 The highly challenging task of training classifiers on imbalanced datasets is a long-standing problem in supervised learning. In real-world applications, like fraud detection, medical diagnosis, and rare event prediction, the class distributions are often highly skewed, leading standard learning algorithms to be heavily biased towards the majority class.
To mitigate this problem, many techniques have been proposed. Existing approaches include resampling-based methods, which either resample statically before the training process or dynamically while training. Static approaches like Random Over-Sampling (ROS) \citep{Chen2024}, Random Under-Sampling (RUS) \citep{Chen2024}, Synthetic Minority Oversampling Technique (SMOTE) \citep{Chawla_2002}, and nearmiss undersampling \citep{mani2003knn} fail to consider the changing difficulty of identifying a class as training progresses. Current dynamic approaches focus on instance-level hardness and often fail to account for imbalance in data. Other approaches are algorithmic, such as cost-sensitive learning \citep{article2}, Online Hard Example Mining (OHEM) \citep{shrivastava2016trainingregionbasedobjectdetectors}, and focal loss \citep{lin2018focallossdenseobject}, which modify the loss function to assign higher costs to misclassifying or minority class examples, thereby increasing the gradient contribution from these instances during optimization. More importantly, these methods fail to consider the changing difficulty of identifying a class as training progresses. While being effective in some cases, all of these methods share common assumptions, as they either (i) treat imbalance as a static property, fixed before the training begins, (ii) focus on instance-level hardness rather than class-level, or (iii) rely on complex heuristics to assign class-specific loss weights. These assumptions ignore the fact that class difficulty evolves during learning \citep{Sinha_2022, toneva2019empiricalstudyexampleforgetting}. A class that initially seems difficult due to low representation may, over time, become easier to learn as the model adapts. Static correction cannot adapt to these dynamics, and often leads to overfitting on the minority class or wasting capacity by oversampling far beyond what is actually needed.
\begin{figure}
    \centering
    \includegraphics[width=1\linewidth]{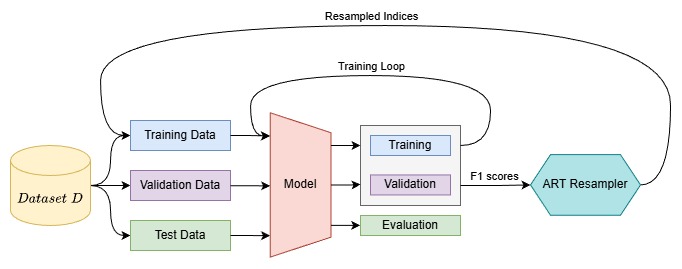}
    \caption{ART resampling }
    \label{fig:overallsys}
\end{figure}
To this end, we propose ART, a resampling-based method that uses the class-wise macro F1-score as a simple heuristic for adjusting the data distribution. This is a novel approach that dynamically resamples data based on model performance. ART periodically monitors class-wise performance and updates the sampling distribution to emphasize underperforming classes. This class-level focus avoids the instability associated with instance-level difficulty estimation \citep{Sinha_2022} and treats class difficulty as dynamic rather than static throughout training. Figure\ref{fig:overallsys} shows how the sampling distribution of classes is changed based on the macro F1-score as training progresses. Using a class-wise metric as a resampling heuristic offers a simpler and more robust alternative to complex fixed heuristics.
 Unlike ~\citep{Sinha_2022}, which uses both resampling and a modified loss function, ART demonstrates that performance metric-based resampling alone outperforms existing methods. Also, ART utilizes a balance of performance metric-based sampling and the original dataset priors rather than completely ignoring the original data distribution. We demonstrate that ART consistently outperforms existing methods across classification tasks, achieving superior accuracy, recall, precision, and F1-scores. We also show that ART performs better than the baseline at varying levels of imbalance. Furthermore, we evaluate ART on classification tasks across tabular, image, and text datasets to demonstrate its modality-agnostic nature.

Unlike existing imbalance-handling approaches that rely on fixed resampling heuristics, instance-level difficulty estimation, or loss reweighting, ART introduces a metric-driven, class-level adaptive resampling strategy. By using periodically evaluated class-wise macro F1  scores as feedback, ART dynamically reallocates training focus based on evolving class difficulty rather than static class priors. This design makes ART model-agnostic, stable, and easy to integrate into standard training pipelines, while directly aligning training adaptation with commonly used imbalance evaluation metrics.

\section{Related Work}

\textbf{Resampling-based methods:} A common approach to handle class imbalance is to rebalance the data distribution before or during training. Classic techniques include ROS, which simply duplicates minority-class examples, and RUS, which deletes majority-class examples. ROS and RUS are simple resampling methods and have been shown to be surprisingly competitive on some tasks. More sophisticated oversampling methods generate synthetic examples. These include methods like SMOTE \citep{Chawla_2002}, which interpolates new minority examples in the feature space. Many SMOTE variants aim to focus sampling on hard or informative regions. For instance, borderline-SMOTE \citep{10.1007/11538059_91} generates points near class boundaries, and ADASYN \citep{4633969} adjusts sampling density based on local minority density. There are also some recent methods like safe-level-SMOTE \citep{10.1007/978-3-642-01307-2_43} and k-means–SMOTE \citep{Douzas_2018} that combine clustering or noise filtering with synthetic generation. In practice, oversampling can introduce noisy or redundant points, so hybrid methods that combine oversampling with cleaning have been proposed in \citep{article1}. For example, SMOTE-ENN and SMOTE-Tomek, as described in \citep{10.1145/1007730.1007735}, use edited nearest neighbors or Tomek-link removal to prune ambiguous samples after SMOTE. It has been shown that these hybrid methods yield well-defined class clusters and often outperform vanilla undersampling, especially when the minority class is minimal \citep{article1, 10.1145/1007730.1007735}.

In parallel, many modern variants of undersampling select the majority of examples based on distance or density. For example, the nearmiss family of methods removes the majority samples closest to the minority class (or vice versa), and variants like nearmiss-2 pick those majority examples that lie near many minority points \citep{mani2003knn}. Such methods exploit geometric margins between classes. Other approaches cluster the data first and then undersample within clusters to preserve structure \citep{10.1007/11823728_41}.

Unlike traditional resampling techniques that rely on fixed, data-level heuristics applied prior to training, ART dynamically adapts sampling probabilities during training based on evolving class-wise performance.

\textbf{Loss-based methods:} Rather than altering the data distribution, many techniques modify the learning algorithm or loss function. Classic cost-sensitive learning assigns a higher weight to the loss component of minority classes \citep{10.5555/1642194.1642224}. One can weigh the loss inversely by class frequency, or use a pre-specified cost matrix \citep{article2}. However, determining optimal costs can be difficult for complex models. Focal loss \citep{lin2018focallossdenseobject} addresses class imbalance in deep learning by changing the loss contribution of easy (well-classified) examples, thereby encouraging the model to focus on harder instances. Similarly, OHEM  \citep{shrivastava2016trainingregionbasedobjectdetectors} dynamically selects the examples with the highest loss within each mini-batch and focuses training on those “hard” samples rather than the full dataset. Although OHEM originated in computer vision, the idea of mining difficult instances can be applied in any domain.

While cost-sensitive and loss-reweighting methods modify the optimization objective, ART instead adjusts the data sampling process using external performance feedback, making it complementary to loss-based imbalance handling techniques.

More recently proposed loss functions explicitly encode the class imbalance. For example, Label-Distribution-Aware Margin (LDAM) loss learns a per-class margin that is inversely related to class frequency, encouraging larger classification margins for minority classes \citep{cao2019learningimbalanceddatasetslabeldistributionaware}. Furthermore, it was shown that when LDAM is combined with a Deferred Reweighting (DRW) schedule, it outperforms standard reweighting on long-tailed image benchmarks. Their method first trains with a margin-aware loss (but without class reweighting), then fine-tunes with inverse-frequency weighting. The idea is to avoid early overfitting to minorities and give a balanced starting point before rebalancing later. Other related approaches include the “class-balanced loss” \citep{Cui_2019_CVPR}, which weights cross-entropy by the effective number of samples per class, and more recent margin-based softmax \citep{ren2020balancedmetasoftmaxlongtailedvisual} or contrastive losses \citep{robinson2021contrastivelearninghardnegative} tailored for imbalance. In general, these loss-based methods act at the instance level. Importantly, many of the above are static in their reweighting schedule: the class weights or sampling target is set a priori or according to a fixed curriculum, rather than being continuously updated based on model performance.

In contrast to margin-based approaches that enforce static class-dependent decision boundaries, ART adapts training emphasis dynamically according to observed class-level performance rather than predefined margins.

\textbf{Dynamic, instance-level and curriculum methods} A recent trend is to make sampling or weighting adaptive during training. Inspired by curriculum learning, some methods gradually shift the sampling distribution or loss focus over the course of learning. For example, Dynamic Curriculum Learning (DCL) \citep{wang2019dynamiccurriculumlearningimbalanced} maintains two schedulers: a sampling scheduler and a loss scheduler. The sampling scheduler starts with an imbalanced-target distribution in early epochs (favoring the majority class) and gradually moves towards a balanced distribution. This is motivated by the observation that training too balanced too early can hurt majority-class representation. DCL shows that such a dynamic schedule yields a better trade-off between overall accuracy and balanced accuracy. Other curriculum-based methods dynamically pick which classes or instances to emphasize. For example, the model might first learn from “easy” majority examples and only later focus on the minority class (or vice versa) \citep{10.1145/1553374.1553380}.

Techniques like self-paced learning \citep{NIPS2010_e57c6b95} or focal loss adjust selection based on how confidently the model predicts each sample. Some recent works, like AdaSampling frameworks \citep{zhang2019adasampleadaptivesamplinghard}, iteratively update example weights by model feedback, although many such methods focus on label-noise scenarios. 

Unlike curriculum and self-paced learning approaches that rely on instance-level difficulty estimation or loss thresholds, ART operates at the class level and uses evaluation-aligned metrics, avoiding noisy instance-wise difficulty signals.

Margin-based sampling methods, like MSYN \citep{inproceedings}, explicitly choose instances near class boundaries to improve learning when the classifier has small margins on some samples. In general, instance-level adaptive sampling contrasts with class-level rebalancing because it can upweight or downweight individual examples even within the same class. Whereas instance-level difficulty methods focus on identifying and emphasizing individual hard samples, ART aggregates difficulty at the class level, resulting in more stable and interpretable adaptation in imbalanced settings.

\textbf{Multi-modal Imbalanced Learning:} In practice, imbalanced methods are applied across domains (vision, NLP, tabular data, etc.). Techniques like random sampling and SMOTE assume only vector inputs and have been used extensively in tabular and image datasets. Modern deep-learning loss reweighting (e.g., LDAM+DRW, focal loss) was popularized on vision tasks but extends naturally to NLP or other modalities. For instance, focal loss has been applied in text classification to handle rare classes, and cost-sensitive learning is common in medical diagnosis or fraud detection on tabular data. Class-balanced sampling and ensemble methods like Balanced Meta-Softmax \citep{ren2020balancedmetasoftmaxlongtailedvisual} and LDAM+DRW have been evaluated on both image benchmarks and real-world text or tabular tasks. Moreover, some recent works explicitly benchmark multiple modalities: for example, Cui et al. \citep{Cui_2019_CVPR} and Cao et al. \citep{cao2019learningimbalanceddatasetslabeldistributionaware} validated their class-balanced and margin-based losses on both image recognition and text classification tasks.
Although prior dynamic strategies adjust weights or sampling schedules heuristically, ART uniquely employs class-wise macro F1  as an explicit feedback signal, directly aligning training adaptation with imbalance evaluation metrics.

Overall, existing deep imbalanced learning methods primarily focus on loss design or instance-level adaptation, whereas ART introduces a simple, model-agnostic class-level feedback mechanism that can be integrated into standard training pipelines.

\section{Proposed Adaptive Resampling-based Training (ART) module}
\begin{figure}[!ht]
  \centering
  \includegraphics[width=0.95\linewidth]{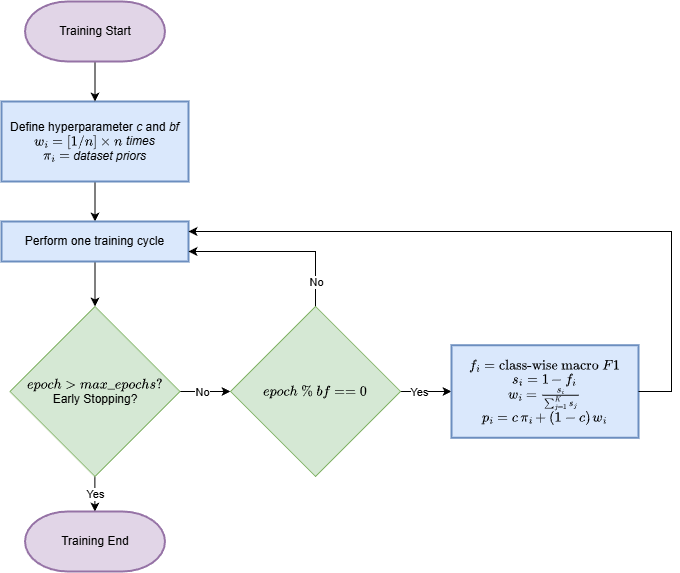}
  \caption{Overview of the training procedure in ART. The process begins with uniform initialization of class weights. Training proceeds in epochs, with periodic updates to class weights every bf epochs based on per-class macro F1-scores evaluated on a validation set.}
  \label{fig:art-dynamics}
\end{figure}

Figure ~\ref{fig:art-dynamics} shows how training progresses under ART.

Consider a training dataset $\mathcal{D} = \{(x_n, y_n)\}_{n=1}^{N}$ containing $N$ labeled samples. Each sample consists of an input feature vector $x_n$ from an input space $\mathcal{X}$ and an associated class label $y_n$ from a set of $K$ classes. Formally:
\begin{align*}
x_n &\in \mathcal{X} \\
y_n &\in \{0, 1, \dots, K-1\}
\end{align*}

We explicitly assume the dataset exhibits significant imbalance across classes, meaning that the distribution of samples is heavily skewed, with some classes represented more frequently than others. The extent of imbalance can adversely affect predictive performance, particularly for minority classes. To characterize this imbalance quantitatively, we define the empirical class prior $\Pi$ for each class $i$ as the relative frequency of that class within the dataset:

\[
\Pi_i = \frac{1}{N} \sum_{n=1}^N \mathbf{1}[y_n = i],
\]

where $N$ is the total number of samples and $\mathbf{1}[\cdot]$ is an indicator function.

Because no difficulty signal is available in the beginning, we set the class weights to a uniform vector,
\[
w^{(0)}_i = \frac{1}{K}, \qquad i = 0,\dotsc,K-1.
\]
 Algorithm~\ref{alg:art} illustrates how these initial weights are blended with the empirical priors via convex interpolation to form the first sampling distribution. The classification model used in this study is a parameterized function, defined by:

\[
f_\theta : \mathcal{X} \rightarrow \Delta^{K-1},
\]

where $\Delta^{K-1}$ denotes the $(K-1)$-simplex, representing valid categorical probability distributions over the $K$ classes. The model parameters, collectively denoted as $\theta$, are optimized by minimizing the expected classification loss under a dynamically adapted sampling distribution $p \in \Delta^{K-1}$. Formally, the optimization problem is given by:

\begin{displaymath}
\hat\theta = \arg\min_{\theta}
  \sum_{i=0}^{K-1} p_i \,
  \mathbb{E}_{(x,y)\sim D_i}\bigl[L\bigl(f_\theta(x),y\bigr)\bigr]
\label{eq:art-objective}
\end{displaymath}

where $D_i \subset D$ denotes the subset of training samples belonging to class $i$, and $L$ represents a suitable classification loss function (e.g., cross-entropy loss).

\subsection{Overview}

The aim of ART is to improve model performance on classification tasks by addressing variations in predictive performance across classes. ART accomplishes this by using an adaptive sampling mechanism informed by the classifier's current performance, rather than relying solely on fixed empirical class frequencies. In contrast to conventional static resampling methods, ART periodically revises the sampling distribution by assessing and quantifying the classification difficulty of each class. 

The algorithm for training and resampling under ART is explained in Algorithm \ref{alg:art}. Classes identified as challenging, indicated by lower predictive performance, receive increased sampling priority. Consequently, ART systematically reallocates training resources towards improving performance on classes exhibiting greater classification complexity.

This dynamic refinement of the sampling distribution occurs iteratively at configurable intervals during the training process. The two hyperparameters that control the behavior of this adaptive procedure are:

\begin{enumerate}
\item the \emph{blending constant}, $c \in [0,1]$, controls the balance between sampling based on empirical class priors and the adaptive distribution derived from current class performance metrics;
\item the \emph{boosting frequency}, $bf \in \mathbb{Z}^+$, specifies the interval, measured in epochs, at which the sampling distribution is updated.
\end{enumerate}

These hyperparameters influence how ART adjusts its sampling strategy over the course of training and affects both the frequency and intensity of its response to class-level learning difficulty changes.

\begin{algorithm}[t]
\caption{Adaptive Resampling–based Training (ART)}
\label{alg:art}
\KwIn{Training set $\mathcal{D}=\{(x_n,y_n)\}_{n=1}^N$, empirical prior $\Pi$, blending factor $c\!\in\![0,1]$, boost frequency $bf\!\in\!\mathbb{Z}^{+}$, total epochs $E$}
\KwOut{Trained model parameters $\hat\theta$}

\textbf{Initialisation}\\
\Indp Initialise model $f_\theta$ (parameters $\theta$)\;
Set \emph{uniform initial class weights}: $w_i \leftarrow \frac{1}{K},\; i=0,\dotsc,K-1$\;
Set \emph{initial sampling distribution}: $p_i \leftarrow c\,\Pi_i + (1-c)\,w_i$\;
\Indm

\For{$\textit{epoch}\leftarrow 1$ \KwTo $E$}{
  Draw minibatches from $\mathcal{D}$ according to $p$\;
  Update $\theta$ by gradient descent on the minibatch loss\;
  \If{$\textit{epoch}\bmod bf = 0$}{
    Evaluate per-class F$1$ scores $f_i$ on the validation set\;
    Compute difficulties $s_i \leftarrow 1 - f_i$\;
    Normalise to new adaptive weights $w_i \leftarrow \dfrac{s_i}{\sum_{j=0}^{K-1}s_j}$\;
    Update sampling distribution $p_i \leftarrow c\,\Pi_i + (1-c)\,w_i$\;
  }
}
\Return{$\hat\theta \leftarrow \theta$}
\end{algorithm}

\subsection{Theoretical Motivation and Optimization Perspective of ART}

ART can be interpreted as a dynamic reweighting scheme that operates directly on the data distribution rather than on the loss function. From an optimization perspective, standard empirical risk minimization (ERM) on imbalanced data minimizes a weighted sum of per-class risks, where the weights are implicitly determined by the empirical class priors. As a result, majority classes dominate the optimization objective, while minority classes contribute weakly to the gradient updates, even when their predictive performance remains poor.

Formally, training on the original dataset corresponds to minimizing the following empirical objective:
\[
\mathcal{R}(\theta) =
\sum_{i=0}^{K-1} \Pi_i \,
\mathbb{E}_{(x,y)\sim D_i}
\left[\mathcal{L}(f_\theta(x), y)\right],
\]
where $\Pi_i$ denotes the empirical prior of class $i$. This objective is static:
The contribution of each class to the optimization process is fixed throughout training, regardless of how class-wise performance evolves. This often results in situations where some classes remain under-optimized even after convergence, especially in severely imbalanced settings.

ART generalizes this by replacing the fixed class prior $\Pi_i$ with a time-varying sampling distribution $p_i^{(t)}$, which is updated based on the model’s observed class-wise performance. 

At the training step $t$, ART approximately optimizes the given objective:
\[
\mathcal{R}^{(t)}(\theta) =
\sum_{i=0}^{K-1} p_i^{(t)} \,
\mathbb{E}_{(x,y)\sim D_i}
\left[\mathcal{L}(f_\theta(x), y)\right].
\]
This highlights that ART performs dynamic ERM, adjusting the relative importance of each class in response to the model’s current deficiencies. Unlike loss-based reweighting methods, which modify gradient magnitudes directly, ART achieves this reweighting indirectly by altering the frequency with which samples from each class are observed during training.

The update rule for $p_i^{(t)}$ is driven by class-wise macro F1-scores evaluated on a held-out validation set. The macro F1-score provides a balanced measure of class-level predictive quality by jointly capturing both precision and recall. Importantly, macro F1 is insensitive to class frequency, ensuring that poor performance on a minority class is not obscured by its limited representation. A low macro F1-score for a class indicates that the model is 
\begin{itemize}
    \item either failing to correctly identify instances of that class (low recall), 
    \item misclassifying many predictions into that class (low precision), or 
    \item both these cases,
\end{itemize}
making it a reliable indicator of class-level learning difficulty.

By defining the difficulty score as $s_i = 1 - f_i$, ART establishes a monotonic relationship between performance and sampling priority: classes with lower predictive performance receive higher sampling probability in subsequent training phases. This design aligns the resampling mechanism with the evaluation objective, ensuring that optimization effort is concentrated on the classes that most negatively impact the macro-level performance metric. From this perspective, ART can be viewed as a feedback-driven training procedure, in which validation-time performance signals guide future data exposure.
\subsection{Performance-based Sampling}

The ART framework periodically evaluates the current predictive performance on each class after every $bf$ epochs. Specifically, the algorithm computes the F1-score for each class $i$ on a held-out validation set $\mathcal{V}$. The class-wise difficulty score, representing how challenging each class is for the current model state, is then defined as:

\[
s_i = 1 - f_i,
\]

where $f_i$ is the class-wise F1-score obtained from the validation set.

These computed difficulty scores reflect the inverse performance relationship and hence, emphasize classes with poor predictive performance. To create a valid probability distribution from these difficulty scores, we normalize them across all classes, resulting in performance-based sampling weights given by:

\[
w_i = \frac{s_i}{\sum_{j=1}^{K} s_j}, 
\quad i = 1, \dots, K.
\]

This normalization guarantees that classes with lower predictive performance will have proportionally higher sampling probabilities, thereby allowing the training algorithm to put greater emphasis towards classes that are currently more difficult to learn. As the model improves on previously challenging classes, the sampling distribution adjusts dynamically, thereby redistributing the focus towards other underperforming classes.

\subsection{Blending with Class Priors}

To ensure stable training, ART does not rely entirely on performance-based adaptive sampling distributions. Instead, it creates a blended sampling distribution by performing a convex interpolation between the empirical class prior distribution $\Pi_i$ and the adaptive performance-based distribution $w_i$. Formally, the blended sampling probability for each class $i$ is defined by:

\[
p_i = c\,\Pi_i + (1 - c)\,w_i,\quad i = 1,\dots,K,
\]

where the blending factor $c \in [0,1]$ explicitly controls the relative weighting between the empirical prior and adaptive distributions. A higher value of $c$ prioritizes sampling according to the original empirical class priors, and a lower value of $c$ prioritizes the adaptive distribution by shifting sampling more towards the currently underperforming classes.

Our convex combination strategy is essential, particularly during the early stages of training, as some classes may temporarily exhibit near-zero performance-based weights ($w_i \approx 0$). Without blending, these classes risk temporary exclusion from the training process, leading to unintended behavior during loss computation. 

\subsection{Computational Overhead}
ART differs from the baseline only in the periodic \emph{refresh} step that updates the sampling distribution and rebuilds the training loader. The baseline trains on a fixed training set and does not perform any refresh. Let $E$ denote the number of epochs actually executed and let $bf$ be the boost frequency. ART triggers a refresh every $bf$ epochs, so the number of refresh events is
\[
R = \floor[\Big]{\frac{E}{bf}}.
\]
We express the total runtime as
\[
T_{ART} = T_{baseline} + \floor[\Big]{\frac{E}{bf}} \cdot T_{refresh},
\]
where $T_{baseline}$ is the baseline wall-clock time for training and validation over $E$ epochs, and $T_{refresh}$ is the wall-clock time of one refresh event.

A refresh event consists of three  components:
\[
T_{refresh} = T_{val\_fwd} + T_{metric} + T_{resample}.
\]

Here,
\begin{itemize}
    \item $T_{val\_fwd}$ is the time for a single forward pass over the validation set.
    \item $T_{metric}$ is the time to compute per-class F1 scores from the validation predictions and calculate the class weights.
    \item $T_{resample}$ is the time to select the indices for resampling based on the class weights.
\end{itemize}

\paragraph{Space overhead}
ART maintains the same asymptotic dataset size as the baseline. It allocates new arrays when rebuilding the refreshed dataset, so peak memory usage may increase during a refresh, but no state proportional to the number of epochs is stored. 

\section{Experiments}

\subsection{Experimental Setup}

We evaluate ART against a broad suite of existing methods across five datasets commonly used for imbalanced classification tasks: Pima Indians Diabetes \citep{smith1988pima}, Yeast \citep{yeast_110}, Red Wine Quality \citep{cortez2009wine}, a custom long-tailed version of MNIST \citep{lecun1998mnist} (MNIST-LT), and a synthetically imbalanced IMDb sentiment classification dataset \citep{maas2011imdb} (IMDb-Custom). MNIST-LT was prepared with an imbalance factor of 0.01, and the number of samples for the majority class was set to 1500. All samples were randomly selected according to the imbalance factor. IMDb-Custom dataset was prepared by transforming the reviews into embedding vectors using all-MiniLM-L6-v2 \citep{allminilm_l6_v2_2021} by sentence-transformers. Then an arbitrary imbalance ratio of $7$ was chosen and samples were randomly chosen from the data in accordance to it. Pima Indians Diabetes dataset is a binary classification dataset with an imbalance ratio of 1.86. Red Wine Quality dataset is a multi-class classification dataset with 6 classes. It has an imbalance ratio ranging from 1.06 to 68.1 in comparison to the class with the most instances. The yeast dataset is a multi-class classification dataset with 9 classes, with imbalance ratios ranging from 1.08 to 23.15 in comparison to the class with the highest number of instances. These datasets encompass both binary and multi-class classification scenarios and vary substantially in terms of class imbalance, feature dimensionality, and input modality (tabular, image, and text).

Each dataset is split into 70\% for training, 15\% for validation, and 15\% for testing. Continuous features in the tabular datasets are standardized using Z-Score Normalization. Bayesian hyperparameter optimization is performed for each method using Optuna’s TPE sampler \citep{akiba2019optuna}. For tabular datasets, we conduct $25 \times n$ trials, where $n$ is the number of hyperparameters (capped at 4 using $n =$ min(no. of hyperparameters, $4)$). For MNIST-LT and IMDb-Custom, we use $5 \times n$ trials due to the computational constraints. All models are trained using the AdamW optimizer \citep{loshchilov2019decoupledweightdecayregularization} with a cosine annealing learning rate scheduler \citep{loshchilov2017sgdrstochasticgradientdescent}. Early stopping is applied with a patience of 10 epochs, based on the loss on the validation set.

We use consistent model architectures across experiments: a lightweight feedforward network for tabular data and IMDb-Custom, and a compact CNN for image classification. Since nearmiss has multiple variations, we use the best variation for each dataset based on hyperparameter optimization. Model-specific configurations are held constant across all baselines to ensure fair comparison.

\subsubsection{Macro F1 as an Appropriate Control Signal for ART}

Choosing the right evaluation metric for adaptive resampling is important because it affects how training resources are distributed among classes. In ART, we use class-wise macro F1 to adjust the sampling distribution for both practical and methodological reasons.

First, macro F1 measures the balance between precision and recall for each class. In imbalanced classification, minority classes often have more false negatives, which lowers recall but does not affect overall accuracy much. Unlike accuracy or cross-entropy loss, class-wise F1 directly penalizes these errors, making it better for improving the recognition of minority classes. Because macro F1 treats all classes equally, it helps identify underperforming classes regardless of how often they appear.

Second, using macro F1 as feedback aligns with the adaptive sampling strategy and the evaluation metric commonly used in imbalanced learning. Many real-world tasks, such as medical diagnosis, fraud detection, and rare-event prediction, prioritize balanced performance over overall accuracy. By using class-wise F1, ART ensures the resampling process uses the same metric used to evaluate model performance, reducing any mismatch between training and evaluation.

Third, using metric-based feedback is more stable than relying on instance-level difficulty. Signals based on loss values or margins can be affected by outliers, noisy labels, and random batch effects. In contrast, class-wise F1 is calculated over the whole validation set, giving a more reliable estimate of class difficulty. This approach helps avoid overreacting to single hard or mislabeled examples, which is a common problem with instance-level sampling.

Finally, macro F1 is easy to use and works with any model. It can be calculated for any classifier and does not depend on the loss function or model structure. This means ART can be used with different types of models and data, as shown in our experiments with tabular, image, and text datasets.

In summary, macro F1 provides a stable, evaluation-focused signal for adaptive resampling. By using this metric at the class level, ART can focus training on classes that need improvement and avoid the instability and complexity of instance-level difficulty modeling.

\subsection{Compared Methods}

We benchmark ART against a range of resampling-based and algorithm-level methods for handling class imbalance:

\paragraph{No Imbalance Handling:}
\begin{itemize}
\item Baseline
\end{itemize}

\paragraph{Resampling-based Methods:}
\begin{itemize}
\item Random Oversampling (ROS)
\item Random Undersampling (RUS)
\item Synthetic Minority Over-sampling Technique (SMOTE)
\item MSMOTE
\item Nearmiss Undersampling
\end{itemize}

\paragraph{Loss-based Methods:}
\begin{itemize}
\item Cost-Sensitive Learning
\item Focal Loss
\item Online Hard Example Mining (OHEM)
\item Label-Distribution-Aware Margin + Deferred Re-Weighting (LDAM + DRW)
\end{itemize}

\paragraph{Hybrid Methods:}
\begin{itemize}
    \item Balanced Meta-Softmax (BALMS)
\end{itemize}

\subsection{Evaluation Protocol}

Each method is evaluated using 20 random seeds to ensure robustness. We report the following:

\begin{itemize}
\item Mean and standard deviation of macro F1 scores on the held-out test set.
\item Paired $t$-test and Wilcoxon signed-rank test to assess the statistical significance of ART compared to each baseline.
\item Average rank of each method across the 20 runs.
\end{itemize}
More information about measures taken to ensure fairness and reproducibility in results can be found in \ref{sec:reproducibility}

\subsection{Test Results}

Table~\ref{tab:results} presents the macro F1 scores across all five datasets.

\begin{table}[t]
  \centering
  \caption{Performance comparison of ART and existing methods on macro F1 score (mean $\pm$ std over 20 seeds).}
  \label{tab:results}
  \resizebox{\textwidth}{!}{
  \begin{tabular}{lccccc}
    \toprule
    \textbf{Method} & \textbf{Pima} & \textbf{Yeast} & \textbf{Red Wine} & \textbf{MNIST-LT} & \textbf{IMDb-Custom} \\
    \midrule

    Baseline & 0.7371 $\pm$ 0.0502 & 0.5152 $\pm$ 0.0531 & 0.3126 $\pm$ 0.0533 & 0.8736 $\pm$ 0.0470 & 0.7258 $\pm$ 0.0171 \\
    \midrule

    \multicolumn{6}{l}{\textbf{Resampling Based Methods}} \\
    \midrule
    ROS & 0.7483 $\pm$ 0.0457 & 0.4727 $\pm$ 0.0419 & 0.3303 $\pm$ 0.0508 & 0.8792 $\pm$ 0.0407 & 0.7166 $\pm$ 0.0211 \\
    RUS & 0.7529 $\pm$ 0.0441 & 0.4286 $\pm$ 0.0501 & 0.1928 $\pm$ 0.0368 & 0.6714 $\pm$ 0.0368 & 0.6839 $\pm$ 0.0214 \\
    SMOTE & 0.7500 $\pm$ 0.0443 & 0.4829 $\pm$ 0.0408 & 0.3281 $\pm$ 0.0487 & --- & --- \\
    MSMOTE & 0.7479 $\pm$ 0.0482 & 0.5008 $\pm$ 0.0417 & 0.2993 $\pm$ 0.0381 & --- & --- \\
    nearmiss & 0.7566 $\pm$ 0.0398 & 0.3767 $\pm$ 0.0521 & 0.1169 $\pm$ 0.0392 & --- & --- \\
    \midrule

    \multicolumn{6}{l}{\textbf{Loss-based Methods}} \\
    \midrule
    Cost-Sensitive Learning & 0.7442 $\pm$ 0.0480 & 0.3912 $\pm$ 0.0689 & 0.2790 $\pm$ 0.0266 & 0.8808 $\pm$ 0.0468 & 0.722 $\pm$ 0.0174 \\
    Focal Loss & 0.7414 $\pm$ 0.0487 & 0.5094 $\pm$ 0.0548 & 0.3193 $\pm$ 0.0512 & 0.8716 $\pm$ 0.0344 & 0.7247 $\pm$ 0.0168 \\
    OHEM & 0.7344 $\pm$ 0.0475 & 0.5007 $\pm$ 0.0380 & 0.2697 $\pm$ 0.0297 & 0.8879 $\pm$ 0.0393 & 0.7197 $\pm$ 0.0171 \\
    LDAM+DRW & 0.7559 $\pm$ 0.0489 & 0.5073 $\pm$ 0.0521 & 0.2878 $\pm$ 0.0246 & 0.8464 $\pm$ 0.0588 & 0.7025 $\pm$ 0.0269 \\
    \midrule

    \multicolumn{6}{l}{\textbf{Hybrid Methods}} \\
    \midrule
    BALMS & 0.7175 $\pm$ 0.0432 & 0.4969 $\pm$ 0.0417 & 0.3263 $\pm$ 0.0404 & 0.8837 $\pm$ 0.0315 & 0.7018 $\pm$ 0.0182 \\
    \midrule

    \textbf{ART (Ours)} & \textbf{0.7631 $\pm$ 0.0498} & \textbf{0.5306 $\pm$ 0.0432} & \textbf{0.3506 $\pm$ 0.0613} & \textbf{0.8848 $\pm$ 0.0359} & \textbf{0.7296 $\pm$ 0.0165} \\
    \bottomrule
  \end{tabular}}
\end{table}

\begin{figure}[!h]
    \centering
    \includegraphics[width=0.95\linewidth]{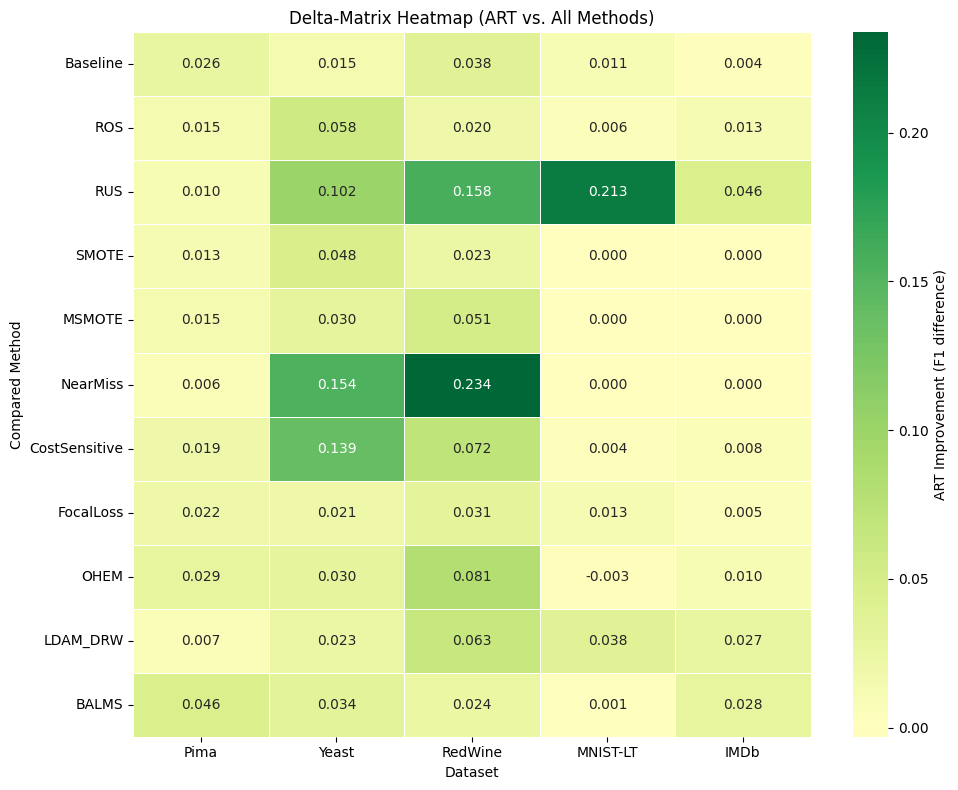}
    \caption{Delta-Matrix heatmap showing the difference in macro F1 scores between ART and competing methods across all datasets. Positive (green) values indicate improvements of ART over the compared method, while negative (red) values denote cases where the competing method performs better.}
    \label{fig:Delta}
\end{figure}

ART achieves the highest macro F1 scores across all datasets, which is visualized in Figure \ref{fig:Delta} using a Delta matrix heatmap that compares ART and competing methods across all datasets.

ART consistently performs well on other metrics, as shown in Table ~\ref{tab:results} and \ref{sec:perf_metrics}. 
In the Pima Indians Diabetes dataset, ART consistently outperforms strong baselines, demonstrating its effectiveness in moderately imbalanced binary classification tasks. The Yeast dataset poses a significant challenge due to its multi-class nature. ART demonstrates clear superiority, likely due to its adaptive class-wise attention mechanism. The Red Wine quality dataset, with its severe class imbalance, emphasizes ART's capacity to handle challenging, skewed distributions effectively. ART maintains stable performance, highlighting its robustness in long-tailed image classification scenarios like the MNIST-LT dataset. It demonstrates competitive performance, illustrating the flexibility of adaptive resampling techniques in text classification of the IMDb-custom dataset.

To verify that ART’s improvements are not due to random seed variation, we test each
baseline against ART with both a paired $t$-test and a two-sided Wilcoxon signed-rank test over the same 20 seeds used in the main experiments. \ref{sec:statistical_tests} reports the resulting p-values, where entries
below 0.05 indicate a statistically significant advantage for ART. We consistently achieve p-value < 0.05 against most methods, indicating significant performance gains when using ART compared to other methods.

\subsection{Ablation Studies}

To evaluate the robustness and sensitivity of ART to its key hyperparameters and design choices, we conducted a series of ablation studies. Due to computational constraints, these experiments were carried out on the Pima Indians Diabetes dataset, which provides a reasonable balance between class imbalance, binary classification, and fast convergence. While the results are specific to this dataset, the trends, particularly ART's stability and consistent performance across different settings, are expected to generalize, given that the method does not rely on any modality-specific assumptions. Future work can expand these studies to include a wider range of datasets and modalities.

\subsubsection{Effect of the Blending Constant \( c \)}

\begin{figure}[h]
  \centering
  \includegraphics[width=0.75\linewidth]{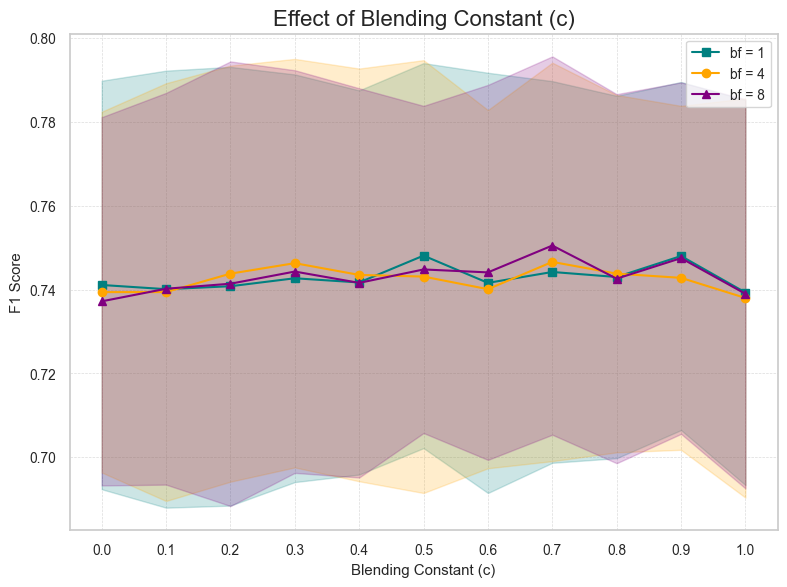}
  \caption{Comparison of ART performance over different blending constant ($c$) where c is a hyperparameter. Results show ART is insensitive to the choice of $c$ and requires minimal hyperparameter optimization.}
  \label{fig:art-ablation-c}
\end{figure}

The blending constant \( c \in [0, 1] \) controls the weight assigned to the static class prior versus the dynamic class-wise performance when creating the sampling distribution. We evaluated ART’s performance using \( c = 0.0, 0.1, 0.2, 0.3, 0.4, 0.5, 0.6, 0.7, 0.8, 0.9, 1.0 \) using three different values of the hyperparameter $bf = 1, 4, 8$. These values cover the low, mid and high range of values of the $bf$ for the dataset used. Figure~\ref{fig:art-ablation-c} shows the value of macro F1-score on these different \(c \) values. Performance remained stable across all \( c \) values, indicating that ART is not sensitive to this parameter. Hence, a mid-range value like \( c = 0.5 \) offers a reliable default setting.

\subsubsection{Effect of Boost Frequency \( bf \)}

\begin{figure}[h]
  \centering
  \includegraphics[width=0.75\linewidth]{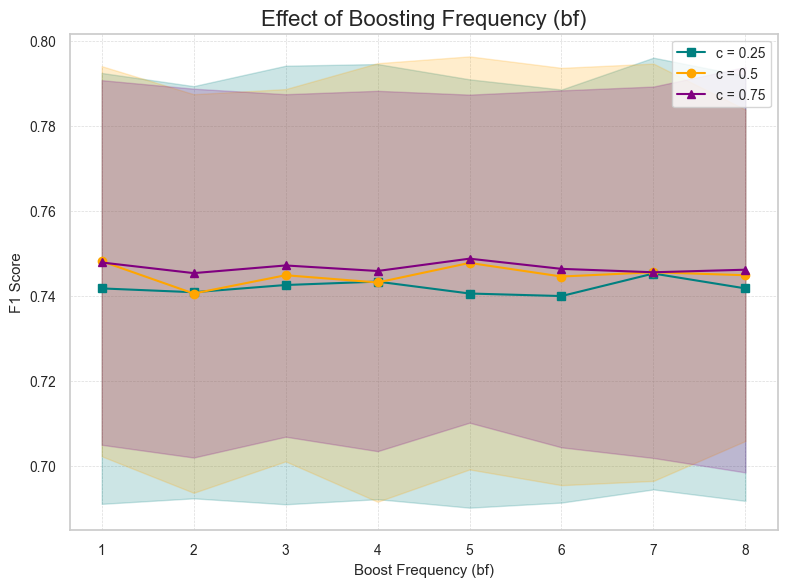}
  \caption{Comparison of performance of ART across varying boosting frequency ($bf$) values while keeping all other hyperparameters constant. Experiments over varying c values show performance is insensitive to choice of boosting frequency.}
  \label{fig:art-ablation-bf}
\end{figure}

The boost frequency \( bf \) controls how often the class distribution is updated during training. We evaluated ART with \( bf = 1, 2, 3, 4, 5, 6, 7, 8, 9, 10 \), covering frequent, moderate, and infrequent updates. Each configuration was tested with three values of the blending constant $c = 0.25, 0.5, 0.75$, covering low, medium and high values. As shown in Figure~\ref{fig:art-ablation-bf}, performance remained consistent across all combinations, suggesting that ART does not require frequent updates to maintain effectiveness. Hence, this reduces the need for extensive hyperparameter tuning in practice.

\subsubsection{Effect of Model Width}

\begin{figure}[h]
  \centering
  \includegraphics[width=0.75\linewidth]{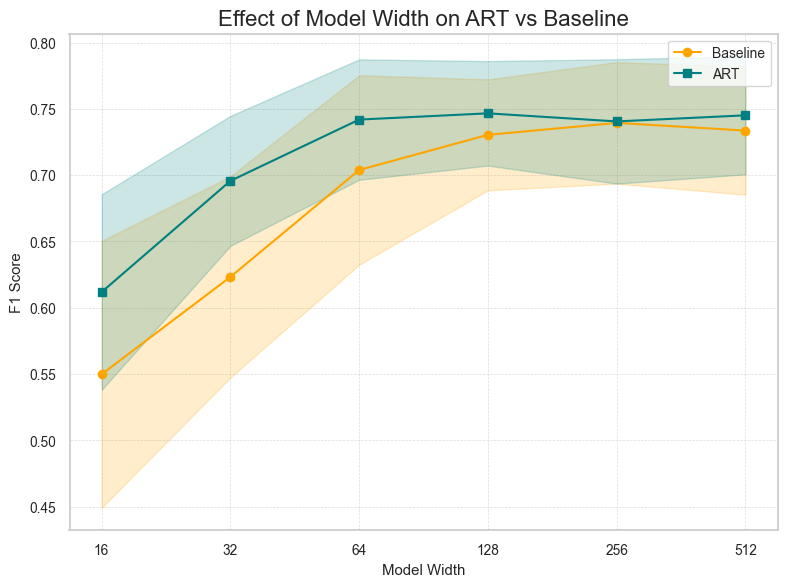}
  \caption{Comparison of baseline and ART performance over different model width. Other model architecture choices and training hyperparameters were kept constant.}
  \label{fig:art-ablation-width}
\end{figure}

To examine ART’s behavior under different model capacities, we varied the initial model width (\( d \)) from 16 to 512 in powers of two while keeping other model architecture choices constant. As shown in Figure~\ref{fig:art-ablation-width}, both ART and the baseline improve with increased width, but ART consistently outperforms the baseline at all sizes. For instance, ART achieves an F1 score of \( 0.7419 \pm 0.0454 \) at width 64, while the baseline requires width 256 to reach a comparable score of \( 0.7394 \pm 0.0458 \). In this case, ART matches performance using a model that is four times smaller. It also exhibits lower standard deviations across widths. This suggests that ART trains more stably across random seeds. It also performs well even when the model is small. This makes ART a good choice for low-resource settings where model capacity is limited.

\subsubsection{Performance Across Imbalance Ratios}

To assess ART's robustness under varying degrees of class imbalance, we generated synthetic variants of the Pima Indians Diabetes dataset with imbalance ratios ranging from 2 to 50. We then compared ART to a baseline model trained without any imbalance mitigation.

\begin{figure}[h]
  \centering
  \includegraphics[width=0.75\linewidth]{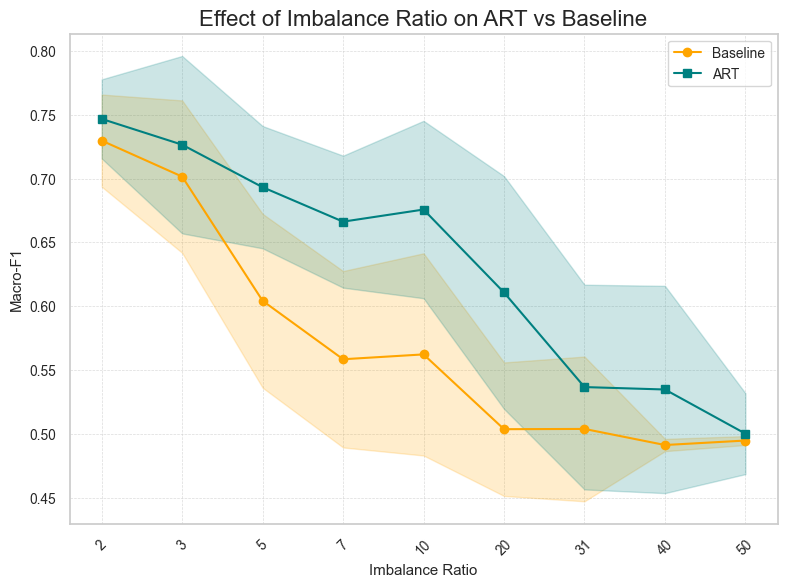}
  \caption{ART maintains higher performance than Baseline at most imbalance levels, with both methods showing a decline as imbalance increases. Shaded areas indicate variability across runs.}
  \label{fig:art-ablation-imb}
\end{figure}

As shown in Figure~\ref{fig:art-ablation-imb}, ART consistently outperforms the baseline across all imbalance ratios. While both methods experience a decline in macro F1 as imbalance severity increases, ART demonstrates significantly better stability. For instance, at an imbalance ratio of 10, ART achieves a macro F1 of approximately 0.68, whereas the baseline drops to around 0.56. Even at extreme imbalance levels, ART maintains a notable advantage, highlighting its ability to adapt sampling in response to increasing class difficulty.

\section{Conclusion}
We conclude by highlighting key results, discussing current limitations, and outlining promising directions for future research.
\subsection{Results}
Across five benchmark datasets covering tabular, image, and text modalities, ART consistently achieved the strongest macro F1 scores, outperforming every method we tested. On the Pima, Yeast, and Red Wine datasets, it improved over the best competitor by 1.0–3.8 percentage points and increased the mean macro-F$1$ across all tabular datasets by 2.64 percentage points. These gains were statistically significant according to paired $t$-tests and Wilcoxon signed-rank tests ($p < 0.05$). ART also maintained its advantage on the long-tailed MNIST-LT and the synthetically imbalanced IMDb-Custom dataset, confirming its modality-agnostic design. Notably, ART matched or exceeded baseline performance even when the underlying model's width was reduced by a factor of four, confirming that it can maintain high performance under limited computational resources.

\subsection{Limitations}
ART assumes an \emph{epoch-based} training loop that (i) computes batch-wise losses, (ii) aggregates gradients, and (iii) evaluates class-wise metrics on a validation set at regular intervals. Thus, it applies well to models trained with stochastic gradient descent (e.g. MLPs, and CNNs) but is unsuitable for non-iterative learners, like decision trees, or k-NN. In addition, computing class-wise F1 every bf epochs incurs overhead proportional to the validation-set size; for very large datasets, this cost may be non-trivial unless subsampling is employed.
ART relies on a held-out validation set to estimate class-wise performance, which may reduce effective training data in extremely small datasets. Additionally, because low F1 scores may arise from label noise rather than intrinsic difficulty, ART may overemphasize noisy classes; mitigating this effect is a direction for future work.

\subsection{Future Work}
Three directions look especially promising:
\begin{enumerate}
\item \textbf{Hybrid sampling.} Trying various methods for dynamic class-level rebalancing in ART could provide better performance as compared to using simple random oversampling and undersampling.
\item \textbf{Broader model classes.} We plan to evaluate ART on large language models and also assess its performance in standard logistic regression pipelines.
\item \textbf{New modalities.} Applying ART to imbalanced audio classification task would test its adaptability to high-rate time-series data where class difficulty can change rapidly during training.
\end{enumerate}
Together, these avenues can deepen our understanding of ART and further widen its applicability in modern machine-learning pipelines.

\section*{Declarations:}
The code is available online at \href{https://github.com/ArjunBasandrai/art}{Github Link}.

\bibliographystyle{model5-names}
\bibliography{reference}

\newpage
\appendix

\section*{Appendix}
\section{Statistical Significance Tests}
\label{sec:statistical_tests}

\begin{table}[h]
\centering
\small
\caption{Pima Dataset: p-values (ART vs. baselines)}
\label{tab:pvals-pima}
\begin{tabular}{lcc}
\toprule
\textbf{Method} & \textbf{Paired t-test} & \textbf{Wilcoxon test} \\
\midrule
Baseline & 0.0001 & 0.0003 \\
ROS & 0.0089 & 0.0141 \\
RUS & 0.0846 & 0.0642 \\
SMOTE & 0.0107 & 0.0023 \\
MSMOTE & 0.0087 & 0.0124 \\
nearmiss & 0.3070 & 0.1327 \\
Cost-Sensitive & 0.0018 & 0.0032 \\
Focal Loss & 0.0014 & 0.0038 \\
OHEM & 0.0005 & 0.0006 \\
LDAM+DRW & 0.1188 & 0.1231 \\
BALMS & 0.0010 & 0.0020 \\
\bottomrule
\end{tabular}
\end{table}

\begin{table}[h]
\centering
\small
\caption{Yeast Dataset: p-values (ART vs. baselines)}
\label{tab:pvals-yeast}
\begin{tabular}{lcc}
\toprule
\textbf{Method} & \textbf{Paired t-test} & \textbf{Wilcoxon test} \\
\midrule
Baseline & 0.1432 & 0.1893 \\
ROS & 0.0000 & 0.0000 \\
RUS & 0.0000 & 0.0000 \\
SMOTE & 0.0000 & 0.0000 \\
MSMOTE & 0.0012 & 0.0025 \\
nearmiss & 0.0000 & 0.0000 \\
Cost-Sensitive & 0.0000 & 0.0000 \\
Focal Loss & 0.0324 & 0.0266 \\
LDAM+DRW & 0.0253 & 0.0266 \\
BALMS & 0.0133 & 0.0136 \\
\bottomrule
\end{tabular}
\end{table}

\begin{table}[h]
\centering
\small
\caption{Red Wine Dataset: p-values (ART vs. baselines)}
\label{tab:pvals-wine}
\begin{tabular}{lcc}
\toprule
\textbf{Method} & \textbf{Paired t-test} & \textbf{Wilcoxon test} \\
\midrule
Baseline & 0.0293 & 0.0153 \\
ROS & 0.2397 & 0.3683 \\
RUS & 0.0000 & 0.0000 \\
SMOTE & 0.1246 & 0.2305 \\
MSMOTE & 0.0081 & 0.0064 \\
nearmiss & 0.0000 & 0.0000 \\
Cost-Sensitive & 0.0003 & 0.0003 \\
Focal Loss & 0.0633 & 0.0583 \\
OHEM & 0.0000 & 0.0000 \\
LDAM+DRW & 0.0003 & 0.0001 \\
BALMS & 0.0717 & 0.1893 \\
\bottomrule
\end{tabular}
\end{table}

\begin{table}[!h]
\centering
\small
\caption{MNIST-LT Dataset: p-values (ART vs. baselines)}
\label{tab:pvals-mnist}
\begin{tabular}{lcc}
\toprule
\textbf{Method} & \textbf{Paired t-test} & \textbf{Wilcoxon test} \\
\midrule
Baseline & 0.1851 & 0.3683 \\
ROS & 0.4923 & 0.5706 \\
RUS & 0.0000 & 0.0000 \\
Cost-Sensitive & 0.7052 & 0.8695 \\
Focal Loss & 0.0874 & 0.0825 \\
LDAM+DRW & 0.0058 & 0.0073 \\
BALMS & 0.8970 & 0.9854 \\
\bottomrule
\end{tabular}
\end{table}

\begin{table}[!h]
\centering
\small
\caption{IMDb-Custom Dataset: p-values (ART vs. baselines)}
\label{tab:pvals-imdb}
\begin{tabular}{lcc}
\toprule
\textbf{Method} & \textbf{Paired t-test} & \textbf{Wilcoxon test} \\
\midrule
Baseline & 0.4482 & 0.3884 \\
ROS & 0.0278 & 0.0318 \\
RUS & 0.0000 & 0.0000 \\
Cost-Sensitive & 0.1714 & 0.1671 \\
Focal Loss & 0.3244 & 0.4330 \\
LDAM+DRW & 0.0001 & 0.0007 \\
BALMS & 0.0000 & 0.0000 \\
\bottomrule
\end{tabular}
\end{table}

\section{Reproducibility Details}
\label{sec:reproducibility}

To facilitate exact replication of our results, we provide below the fixed random-seed list, hardware configuration, and software environment used for every experiment in Section 4. The code is available online at \href{https://github.com/ArjunBasandrai/art}{Github Link}.

\subsection{Random Seeds}

All runs were executed with the following 20 integer seeds, applied uniformly
to \texttt{NumPy}, \texttt{PyTorch}, and (where applicable) \texttt{CUDA}
\texttt{We used the following seeds for running our tests: 1834, 8993, 412, 4523, 182, 41921, 53178, 4536, 89, 101172, 3812, 76459, 21734, 5601, 14923, 32871, 982, 61435, 23490, 7711. Metrics for each method are averaged over all 20 seeds.}

\subsection{Hardware}

\begin{itemize}
  \item \textbf{GPU.} All training was performed on a single NVIDIA Tesla P100 (16 GB) provided by Kaggle notebooks.  
  \item \textbf{CPU.}  1 × Intel Xeon (model 85, Skylake) @ 2.00 GHz, 4 vCPUs  (2 cores × 2 threads), little-endian, AVX-512 capable, 38.5 MiB shared L3.
  \item \textbf{RAM.}  31 GiB system RAM (\texttt{free -g} reports $\approx$22 GiB free and $\approx$7 GiB cached at notebook start).
\end{itemize}
All experiments ran inside a standard Kaggle notebook container

\subsection{Determinism Controls}

Before every run, we invoked the helper below to seed all relevant RNGs and disable non-deterministic cuDNN paths:

\begin{verbatim}
def seed_env(SEED):
    random.seed(SEED)
    np.random.seed(SEED)
    torch.manual_seed(SEED)
    if torch.cuda.is_available():
        torch.cuda.manual_seed(SEED)
        torch.cuda.manual_seed_all(SEED)
    torch.backends.cudnn.deterministic = True
    torch.backends.cudnn.benchmark = False
\end{verbatim}

These details should allow the readers to reproduce the results reported in Tables \ref{tab:results} and \ref{tab:pvals-pima}–\ref{tab:pvals-imdb}.

\section{Performance Metrics}
We demonstrate the performance of ART against existing methods on various metrics. See Table ~\ref{tab:results_accuracy} for performance on accuracy, Table ~\ref{tab:results_precision} for performance on precision and Table ~\ref{tab:results_recall} for performance on recall.
\label{sec:perf_metrics}
\begin{table}[h]
  \centering
  \caption{Performance comparison of ART and existing methods on Accuracy (mean $\pm$ std over 20 seeds).}
  \label{tab:results_accuracy}
  \resizebox{\textwidth}{!}{
  \begin{tabular}{lccccc}
    \toprule
    \textbf{Method} & \textbf{Pima} & \textbf{Yeast} & \textbf{Red Wine} & \textbf{MNIST-LT} & \textbf{IMDb-Custom} \\
    \midrule
    Baseline & 0.7636 $\pm$ 0.0408 & 0.5853 $\pm$ 0.0254 & 0.6100 $\pm$ 0.0261 & 0.9777 $\pm$ 0.0060 & \textbf{0.9218 $\pm$ 0.0061} \\
    ROS & 0.7633 $\pm$ 0.0412 & 0.5396 $\pm$ 0.0377 & 0.5789 $\pm$ 0.0398 & 0.9766 $\pm$ 0.0119 & 0.9157 $\pm$ 0.0084 \\
    RUS & 0.7672 $\pm$ 0.0412 & 0.4536 $\pm$ 0.0473 & 0.2881 $\pm$ 0.0677 & 0.8587 $\pm$ 0.0315 & 0.8491 $\pm$ 0.0215 \\
    SMOTE & 0.7663 $\pm$ 0.0402 & 0.5340 $\pm$ 0.0342 & 0.5527 $\pm$ 0.0408 & --- & --- \\
    MSMOTE & 0.7603 $\pm$ 0.0442 & 0.5835 $\pm$ 0.0268 & 0.5968 $\pm$ 0.0389 & --- & --- \\
    nearmiss & 0.7698 $\pm$ 0.0363 & 0.3804 $\pm$ 0.0577 & 0.1625 $\pm$ 0.0776 & --- & --- \\
    Cost-Sensitive Learning & 0.7573 $\pm$ 0.0443 & 0.4806 $\pm$ 0.0361 & 0.5577 $\pm$ 0.0394 & 0.9726 $\pm$ 0.0139 & 0.9049 $\pm$ 0.0133 \\
    Focal Loss & 0.7668 $\pm$ 0.0402 & 0.5779 $\pm$ 0.0324 & 0.6035 $\pm$ 0.0302 & 0.9766 $\pm$ 0.0063 & 0.9212 $\pm$ 0.0078 \\
    OHEM & 0.7599 $\pm$ 0.0402 & 0.5774 $\pm$ 0.0230 & 0.5862 $\pm$ 0.0373 & \textbf{0.9778 $\pm$ 0.0060} & 0.9214 $\pm$ 0.0061 \\
    LDAM+DRW & 0.7728 $\pm$ 0.0423 & 0.5768 $\pm$ 0.0354 & \textbf{0.6120 $\pm$ 0.0300} & 0.9674 $\pm$ 0.0170 & 0.8884 $\pm$ 0.0239 \\
    \textbf{ART (Ours)} & \textbf{0.7758 $\pm$ 0.0502} & \textbf{0.5914 $\pm$ 0.0332} & 0.5737 $\pm$ 0.0372 & 0.9770 $\pm$ 0.0065 & 0.9083 $\pm$ 0.0183 \\
    \bottomrule
  \end{tabular}}
\end{table}

\begin{table}[h]
  \centering
  \caption{Performance comparison of ART and existing methods on Precision (mean $\pm$ std over 20 seeds).}
  \label{tab:results_precision}
  \resizebox{\textwidth}{!}{
  \begin{tabular}{lccccc}
    \toprule
    \textbf{Method} & \textbf{Pima} & \textbf{Yeast} & \textbf{Red Wine} & \textbf{MNIST-LT} & \textbf{IMDb-Custom} \\
    \midrule
    Baseline & 0.7495 $\pm$ 0.0515 & 0.5565 $\pm$ 0.0732 & 0.3251 $\pm$ 0.0729 & 0.8976 $\pm$ 0.0491 & \textbf{0.7762 $\pm$ 0.0333} \\
    ROS & 0.7482 $\pm$ 0.0478 & 0.4708 $\pm$ 0.0522 & 0.3393 $\pm$ 0.0622 & 0.9062 $\pm$ 0.0468 & 0.7474 $\pm$ 0.0277 \\
    RUS & 0.7544 $\pm$ 0.0493 & 0.4350 $\pm$ 0.0618 & 0.2375 $\pm$ 0.0344 & 0.6606 $\pm$ 0.0324 & 0.6540 $\pm$ 0.0179 \\
    SMOTE & 0.7518 $\pm$ 0.0469 & 0.4794 $\pm$ 0.0545 & 0.3369 $\pm$ 0.0657 & --- & --- \\
    MSMOTE & 0.7478 $\pm$ 0.0496 & 0.5372 $\pm$ 0.0607 & 0.3094 $\pm$ 0.0358 & --- & --- \\
    nearmiss & 0.7566 $\pm$ 0.0427 & 0.3893 $\pm$ 0.0566 & 0.2094 $\pm$ 0.0479 & --- & --- \\
    Cost-Sensitive Learning & 0.7457 $\pm$ 0.0512 & 0.4811 $\pm$ 0.1131 & 0.2879 $\pm$ 0.0302 & 0.8951 $\pm$ 0.0594 & 0.7234 $\pm$ 0.0318 \\
    Focal Loss & 0.7525 $\pm$ 0.0507 & 0.5591 $\pm$ 0.0742 & 0.3312 $\pm$ 0.0725 & 0.9034 $\pm$ 0.0386 & 0.7744 $\pm$ 0.0332 \\
    OHEM & 0.7459 $\pm$ 0.0505 & 0.5436 $\pm$ 0.0484 & 0.2973 $\pm$ 0.0340 & \textbf{0.9134 $\pm$ 0.0465} & 0.7746 $\pm$ 0.0270 \\
    LDAM+DRW & 0.7515 $\pm$ 0.0508 & 0.5490 $\pm$ 0.0597 & 0.3392 $\pm$ 0.0490 & 0.9112 $\pm$ 0.0470 & 0.7522 $\pm$ 0.0236 \\
    \textbf{ART (Ours)} & \textbf{0.7673 $\pm$ 0.0519} & \textbf{0.5644 $\pm$ 0.0595} & \textbf{0.3583 $\pm$ 0.0645} & 0.9068 $\pm$ 0.0437 & 0.7341 $\pm$ 0.0330 \\
    \bottomrule
  \end{tabular}}
\end{table}

\begin{table}[h]
  \centering
  \caption{Performance comparison of ART and existing methods on Recall (mean $\pm$ std over 20 seeds).}
  \label{tab:results_recall}
  \resizebox{\textwidth}{!}{
  \begin{tabular}{lccccc}
    \toprule
    \textbf{Method} & \textbf{Pima} & \textbf{Yeast} & \textbf{Red Wine} & \textbf{MNIST-LT} & \textbf{IMDb-Custom} \\
    \midrule
    Baseline & 0.7331 $\pm$ 0.0496 & 0.5229 $\pm$ 0.0521 & 0.3136 $\pm$ 0.0489 & 0.8643 $\pm$ 0.0475 & 0.6969 $\pm$ 0.0227 \\
    ROS & 0.7530 $\pm$ 0.0451 & 0.5275 $\pm$ 0.0474 & 0.3504 $\pm$ 0.0655 & 0.8706 $\pm$ 0.0356 & 0.6958 $\pm$ 0.0224 \\
    RUS & 0.7584 $\pm$ 0.0416 & 0.4903 $\pm$ 0.0488 & 0.2999 $\pm$ 0.0991 & 0.7714 $\pm$ 0.0510 & \textbf{0.7777 $\pm$ 0.0208} \\
    SMOTE & 0.7532 $\pm$ 0.0425 & \textbf{0.5476 $\pm$ 0.0439} & 0.3619 $\pm$ 0.0707 & --- & --- \\
    MSMOTE & 0.7566 $\pm$ 0.0462 & 0.5126 $\pm$ 0.0375 & 0.3108 $\pm$ 0.0507 & --- & --- \\
    Nearmiss & 0.7635 $\pm$ 0.0386 & 0.4732 $\pm$ 0.0636 & 0.3028 $\pm$ 0.0879 & --- & --- \\
    Cost-Sensitive Learning & 0.7525 $\pm$ 0.0463 & 0.4326 $\pm$ 0.0670 & 0.2972 $\pm$ 0.0265 & \textbf{0.8822 $\pm$ 0.0407} & 0.7318 $\pm$ 0.0357 \\
    Focal Loss & 0.7374 $\pm$ 0.0482 & 0.5176 $\pm$ 0.0499 & 0.3214 $\pm$ 0.0460 & 0.8613 $\pm$ 0.0360 & 0.6964 $\pm$ 0.0234 \\
    OHEM & 0.7314 $\pm$ 0.0470 & 0.5097 $\pm$ 0.0450 & 0.2728 $\pm$ 0.0266 & 0.8778 $\pm$ 0.0388 & 0.6885 $\pm$ 0.0223 \\
    LDAM & 0.7346 $\pm$ 0.0505 & 0.5367 $\pm$ 0.0473 & 0.3250 $\pm$ 0.0424 & 0.8794 $\pm$ 0.0451 & 0.7042 $\pm$ 0.0227 \\
    DRW & 0.7581 $\pm$ 0.0491 & 0.5325 $\pm$ 0.0532 & 0.2908 $\pm$ 0.0223 & 0.8591 $\pm$ 0.0455 & 0.7282 $\pm$ 0.0254 \\
    \textbf{ART (Ours)} & \textbf{0.7703 $\pm$ 0.0428} & 0.5458 $\pm$ 0.0485 & \textbf{0.3783 $\pm$ 0.0767} & 0.8777 $\pm$ 0.0385 & 0.7349 $\pm$ 0.0291 \\
    \bottomrule
  \end{tabular}}
\end{table}

\end{document}